\documentclass[runningheads]{llncs}
\usepackage{graphicx}
\usepackage{comment}	
\usepackage{amsmath,amssymb} %
\usepackage{color}

\usepackage[width=122mm,left=12mm,paperwidth=146mm,height=193mm,top=12mm,paperheight=217mm]{geometry}

\usepackage{pifont}

\usepackage{graphicx}

\usepackage{amsmath}
\usepackage{amssymb}
\usepackage{amsfonts}

\usepackage{array}
\usepackage{booktabs} %
\usepackage{multirow} 
\usepackage{tabu}
\usepackage{tabularx, makecell}
\usepackage[table]{xcolor}
\usepackage{arydshln}

\usepackage{graphicx}
\usepackage{comment}	

\usepackage{color}

\usepackage{enumitem}
\usepackage[normalem]{ulem}

\usepackage{setspace}

\usepackage{bm}

\usepackage{enumerate}

\usepackage{subcaption}

\usepackage{url}  %

\usepackage{float}

\newlength\paramargin
\newlength\figmargin
\newlength\subfigmargin
\newlength\secmargin
\newlength\subsecmargin
\newlength\tabmargin
\newlength\eqmargin

\definecolor{dgreen}{rgb}{0, 0.6, 0} 
\definecolor{cyan}{rgb}{0.88, 1, 1}
\definecolor{bca}{rgb}{0.88, 1, 1}
\definecolor{bcb}{rgb}{0.88, 1, 1}
\definecolor{bcc}{rgb}{0.88, 1, 1}
\definecolor{bcd}{rgb}{0.88, 1, 1}
\definecolor{rca}{rgb}{1, 0.9, 0.9}
\definecolor{rcc}{rgb}{1, 0.9, 0.9}
\definecolor{Darkviolet}{rgb}{0.58, 0, 0.83} 
\definecolor{Darkpink}{rgb}{0.74, 0.2, 0.64}

\definecolor{nice-blue}{HTML}{0071bc}
\definecolor{nice-red}{HTML}{E41A1C}

\newcommand{\etal}{\unskip\ {{et al.}}}
\newcommand{\ie}{{\it i.e.}}

\newcommand{\etc}{{\it etc}}

\newcommand{\heading}[1]{\noindent\textbf{#1}}

\newcommand{\figref}[1]{Fig.~\ref{fig:#1}}%
\newcommand{\tabref}[1]{Table~\ref{tab:#1}} %
\newcommand{\eqnref}[1]{Equation~(\ref{eq:#1})}

\newcommand{\ignore}[1]{}   %

\newcommand{\wzx}[1]{\textcolor{blue}{ZX: #1}}

\newcommand{\hsinyu}[1]{\textcolor{Darkpink}{hsinyu: #1}}
\newcommand{\yulunliu}[1]{{\color{red}\textbf{yulunliu: }#1}\normalfont}

\newcommand{\best}[1]{{\textcolor{red}{#1}}}
\newcommand{\second}[1]{{\textcolor{blue}{#1}}}

\makeatletter
\def\hlinewd#1{%
  \noalign{\ifnum0=`}\fi\hrule \@height #1 \futurelet
   \reserved@a\@xhline}
\makeatother

\makeatletter
\def\adl@drawiv#1#2#3{%
        \hskip.5\tabcolsep
        \xleaders#3{#2.5\@tempdimb #1{1}#2.5\@tempdimb}%
                #2\z@ plus1fil minus1fil\relax
        \hskip.5\tabcolsep}
\newcommand{\cdashlinelr}[1]{%
  \noalign{\vskip\aboverulesep
           \global\let\@dashdrawstore\adl@draw
           \global\let\adl@draw\adl@drawiv}
  \cdashline{#1}
  \noalign{\global\let\adl@draw\@dashdrawstore
           \vskip\belowrulesep}}
\makeatother

\usepackage[pagebackref=false,breaklinks=true,letterpaper=true,colorlinks,citecolor=blue,linkcolor=blue,bookmarks=false]{hyperref}

\usepackage{ifthen}
\newcommand{\final}{0}

\ifthenelse{\equal{\final}{1}}
{
\renewcommand{\wzx}[1]{}
\renewcommand{\hsinyu}[1]{}
\renewcommand{\yulunliu}[1]{}
}{}

\newcommand{\itoi}[1]{{{#1}}}

\definecolor{Gray}{gray}{.95}

\begin{document}
\pagestyle{headings}
\mainmatter
\def\ECCVSubNumber{634}  %

\title{Domain-Specific Mappings\\for Generative Adversarial Style Transfer}

\author{Hsin-Yu Chang \and
Zhixiang Wang \and
Yung-Yu Chuang
}
\institute{National Taiwan University}

\maketitle

\begin{center}
    \centering
    \includegraphics[width=0.99\textwidth]{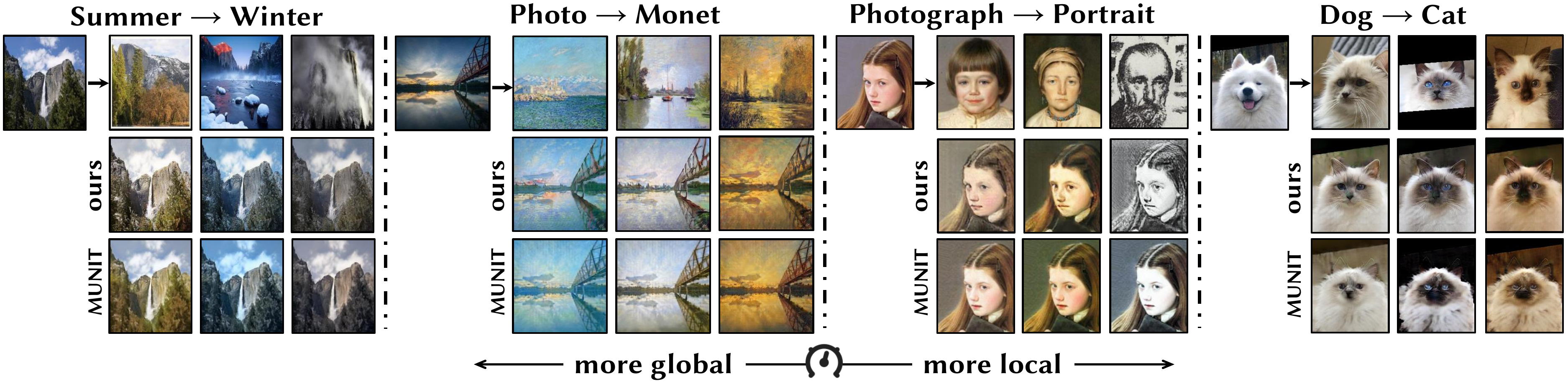}
    \captionof{figure}{
    {\bf Our style transfer results in comparison with MUNIT~\cite{huang2018munit}.} 
    From left to right, style transfer between two domains becomes ``more local'' from ``more global''. For the {Summer}$\to${Winter} task, the style transfer is ``more global'' as its success mainly counts on adjusting global attributes such as color tones and textures. On the contrary, for the {Dog}$\to${Cat} task, style transfer is ``more local'' as its success requires more attention on local and structural semantic correspondences, such as ``eyes to eyes'' and ``nose to nose''. 
    }
  \label{fig:intro_whole}
\end{center}%

\begin{abstract}
Style transfer generates an image whose content comes from one image and style from the other. Image-to-image translation approaches with disentangled representations have been shown effective for style transfer between two image categories. However, previous methods often assume a shared domain-invariant content space, which could compromise the content representation power. For addressing this issue, this paper leverages domain-specific mappings for remapping latent features in the shared content space to domain-specific content spaces. This way, images can be encoded more properly for style transfer. Experiments show that the proposed method outperforms previous style transfer methods, particularly on challenging scenarios that would require semantic correspondences between images. 
Code and results are available at \url{https://acht7111020.github.io/DSMAP-demo/}.

\end{abstract}
\section{Introduction}
\label{sec:intro}

\begin{figure}[t]
  \centering
  \includegraphics[width=.98\textwidth]{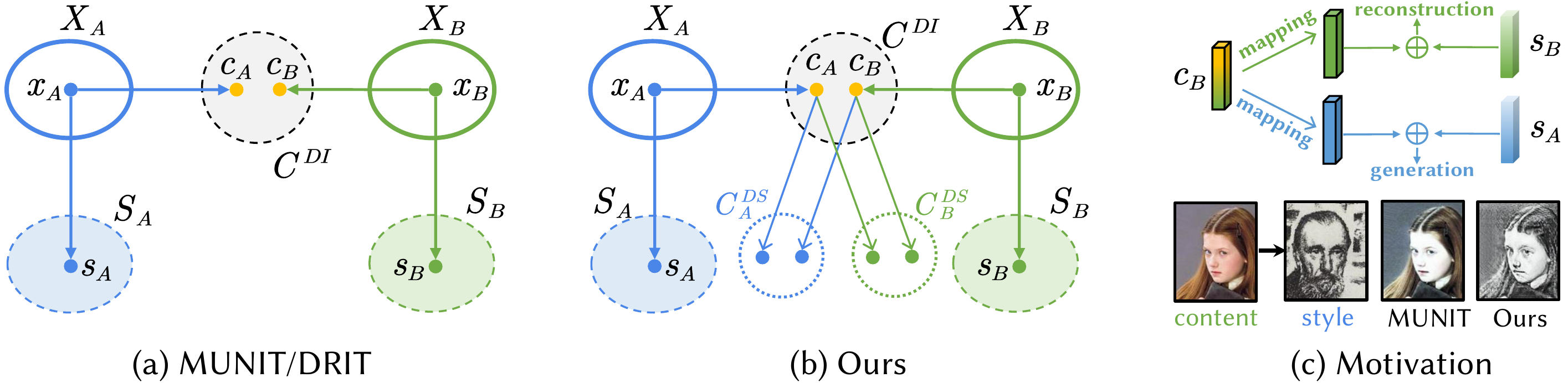}   
  \caption{
  {\bf Main idea and the motivation.}
  (a) MUNIT~\cite{huang2018munit} and DRIT~\cite{lee2018drit} decompose an image into a content feature in the shared domain-invariant content space and a style feature in the domain-specific style space. The shared space could limit representation power. (b) Our method finds the mappings to map a content feature in the shared latent space into each domain's domain-specific content space. This way, the remapped content feature can be better aligned with the characteristics of the target domain. (c) For an image of the source domain, by mapping its content feature to the target domain and combining it with the target domain's style, our method can synthesize better results. %
  } 
  \label{fig:intro_highlevel}
  \vspace{\figmargin}
\end{figure} 

Style transfer has gained great attention recently as it can create interesting and visually pleasing images. It has wide applications such as art creation and image editing. Given a pair of images, the content image $x_A$ and the style image $x_B$, style transfer generates an image with $x_A$'s content and $x_B$'s style. It is not easy to define the content and style precisely. However, in general, the content involves more in the layout and spatial arrangement of an image while the style refers more to the colors, tones, textures, and patterns. \figref{intro_whole} gives examples of our style transfer results in comparison with an existing method, MUNIT~\cite{huang2018munit}.

The seminal work of Gatys \etal~\cite{gatys2016image,gatys2015texture} shows that deep neural networks (DNNs) can extract the correlations (via Gram matrix) of the style and content features, and uses an iterative optimization method for style transfer. Since then, many methods have been proposed to address issues such as generalizing to unseen styles, reducing computation overhead, and improving the matching of the style and content. Image-to-image (I2I) translation aims at learning the mapping between images of two domains (categories) and can be employed for style transfer between two image categories naturally. Recently, some I2I translation approaches have shown great success through disentangled representations, such as MUNIT~\cite{huang2018munit} and DRIT~\cite{lee2018drit}. Although these methods work well for many translation problems, they only give inferior results for some challenging style transfer scenarios, particularly those requiring semantics matches such as eyes to eyes in the Dog-to-Cat transfer.

This paper improves upon the I2I translation approaches with disentangled representations for style transfer. 
We first give an overview of the I2I approaches using disentangled representations as shown in \figref{intro_highlevel}(a). 
Inspired by CycleGAN~\cite{zhu2017cyclegan} which defines two separated spaces and UNIT~\cite{liu2017unsupervised} which assumes a shared latent space, MUNIT~\cite{huang2018munit} and DRIT~\cite{lee2018drit} encode an image $x$ with two feature vectors, the content feature $c$ and the style feature $s$.
In their setting, although domains $X_A$ and $X_B$ have their own latent domain-specific feature spaces for styles $S_A$ and $S_B$, they share the same latent domain-invariant space $C^{DI}$ for the content. Thus, given a content image $x_A\!\!\in\!\!X_A$ and a style image $x_B\!\!\in\!\!X_B$, they are encoded as $(c_A, s_A)$ and $(c_B, s_B)$ respectively. The content features $c_A$ and $c_B$ belong to the shared domain-invariant content space $C^{DI}$ while $s_A$ and $s_B$ respectively belong to the style space of its own domain, $S_A$ and $S_B$. For the cross-domain translation task $A\!\!\rightarrow\!\!B$, the content feature $c_A$ and the style feature $s_B$ are fed into a generator for synthesizing the result with $x_A$'s content and $x_B$'s style. 

\figref{intro_whole} shows MUNIT's results for several style transfer tasks. We found that previous I2I methods with disentangled representations often run into problems in ``more local'' style transfer scenarios. 
We observe that the shared domain-invariant content space could compromise the ability to represent content since they do not consider the relationship between content and style. We conjecture that the domain-invariant content feature may contain domain-related information, which causes problems in style transfer.
To address the issue, we leverage two additional domain-specific mapping functions $\Phi_{C \rightarrow C_A}$ and $\Phi_{C \rightarrow C_B}$ to remap the content features in the shared domain-invariant content space $C^{DI}$ into the features in the domain-specific content spaces $C_A^{DS}$ and $C_B^{DS}$ for different domains (\figref{intro_highlevel}(b)). The domain-specific content space could better encode the domain-related information needed for translation by representing the content better. Also, domain-specific mapping helps the content feature better align with the target domain.
Thus, the proposed method improves the quality of translation and handles both local and global style transfer scenarios well. 

The paper's main contribution is the observation that both the content and style have domain-specific information and the proposal of the domain-specific content features. Along this way, we design the domain-specific content mapping and propose proper losses for its training. The proposed method is simple yet effective and has the potential to be applied to other I2I translation frameworks.

\section{Related work}

\heading{Style transfer.} Gatys~\etal~\cite{gatys2016image} show that the image representation derived from CNNs can capture the style information and propose an iterative optimization method to calculate the losses between the target and input images. 
For reducing the substantial computational cost of the optimization problem, some propose to use a feed-forward network~\cite{johnson2016perceptual,chen2017stylebank}. Although generating some good results, these methods only have limited ability to transfer to an unseen style.

For general style transfer, Huang~\etal~\cite{huang2017adain} propose a novel adaptive instance normalization (AdaIN) layer for better aligning the means and variances of the features between the content and style images. 
The WCT algorithm~\cite{li2017universal} embeds a whitening and coloring process to encode the style covariance matrix. Li~\etal~\cite{li2018learning} propose a method that can reduce the matrix computation by learning the transformation matrix with a feed-forward network.
Although they can generalize to unseen styles, these networks cannot learn the semantic style translation, such as transferring from a cat to a dog. 
Cho~\etal~\cite{cho2019gdwct} propose a module that allows image translation for conveying profound style semantics by using the deep whitening-and-coloring approach.

There are approaches explicitly designed for semantic style transfer. Lu~\etal~\cite{lu2017semantic} propose a method for generating semantic segmentation maps using VGG, while Luan~\etal~\cite{luan2017deep} use the model provided by others for obtaining segmentation masks. 
The success of these methods heavily depends on the quality of the segmentation masks. For scenarios where there is no good correspondence between masks, these methods can not work well. 
To reduce computational overhead, Lu~\etal~\cite{lu2017semantic} decompose the semantic style transfer problem into two sub-problems of feature reconstruction and feature decoding. 
Liao~\cite{liao2017semantic} solve the visual attribute transfer problem when the content and style images have perceptually similar semantic structures. However, it can not work on the images that have less similar semantic structures. 
Our method learns the semantic correspondences implicitly through content space projection. Thus, it works better even if there is no explicit semantic correspondence between images. 

\begin{figure*}[tbp]
  \centering
  \includegraphics[width=1.0\textwidth]{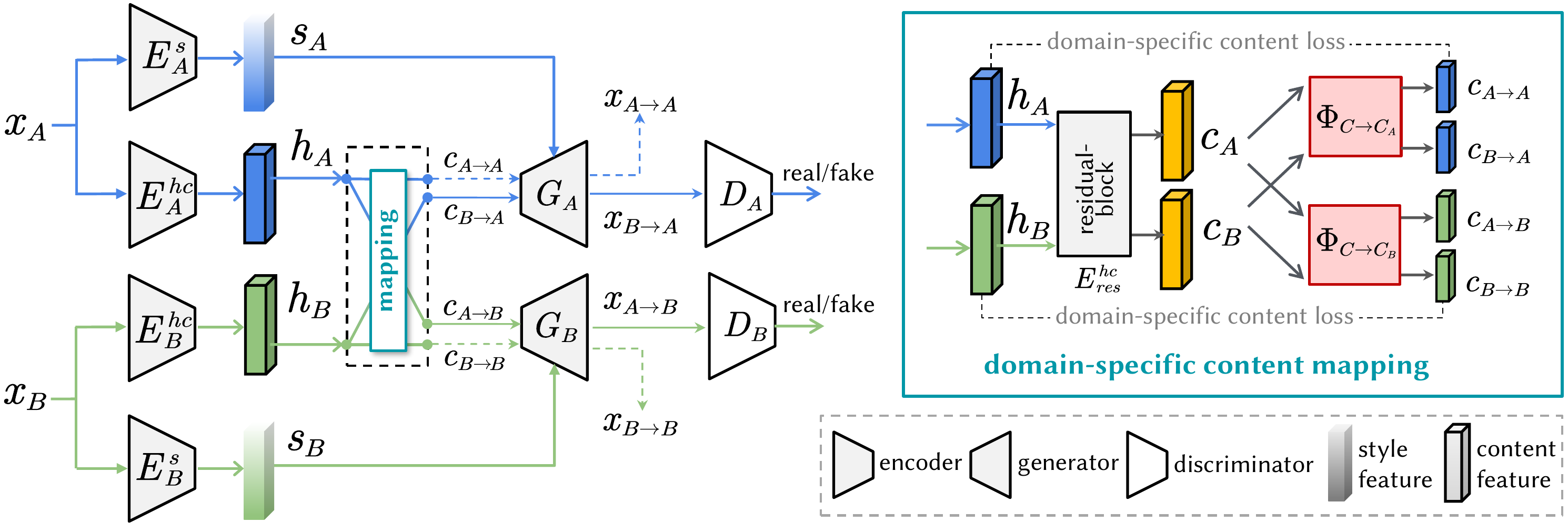}
  \caption{{\bf The framework of the proposed method.}
  For improving cross-domain translation, we use domain-specific mappings for obtaining content features $c_{A\rightarrow{B}}$ and $c_{B\rightarrow{A}}$ instead of domain-invariant content features $c_A$ and $c_B$ for better preserving domain-specific information. The details of the domain-specific content mapping are shown in the box on the top right. 
  }
   \vspace{\figmargin}
  \label{fig:overview}
\end{figure*}

\heading{Image-to-image translation.}
Using the cGAN framework, Pix2Pix~\cite{isola2017pix2pix} models the mapping function from the source domain to the target domain. BicycleGAN~\cite{zhu2017bicyclegan} adds the variational autoencoder (VAE)~\cite{kingma2013vae} in cGAN to generate multiple results from a single image, named as multi-modal translation. 
Both Pix2Pix and BicycleGAN require paired training images. 
For training with unpaired images, CycleGAN~\cite{zhu2017cyclegan} and DualGAN~\cite{yi2017dualgan} employ a novel constraint, the cycle consistency loss, in the unsupervised GAN framework. The idea is to obtain the same image after transferring to the target domain and then transferring back.
Several researchers~\cite{choi2018stargan,pumarola2018ganimation,xiao2018elegant} use the cycle constraint to generate portraits with different facial attributes. 
To maintain the background information, some methods~\cite{mejjati2018unsupervised,pumarola2018ganimation,mo2018instagan,Kim2019UGATIT} add attention masks in their architectures. In addition to facial images, some deal with images of more classes~\cite{mejjati2018unsupervised,mo2018instagan,Kim2019UGATIT}.

UNIT~\cite{liu2017unsupervised} assumes that two domains map to a shared latent space and learns a unimodal mapping. 
To deal with multimodal translation problems, DRIT~\cite{lee2018drit} and MUNIT~\cite{huang2018munit} extend the setting of UNIT~\cite{liu2017unsupervised} and decompose images into a shared content space and different attribute spaces for different domains. To learn individual attribute features, they design a training process for learning the style encoders and the content encoders. However, these methods tend to generate poor style transfer results when there is a large gap between domains. 
Zheng~\etal~\cite{zheng2019multibranch} use a multi-branch discriminator to learn the locations or numbers of objects to be transferred.
TransGaGa~\cite{wu2019transgaga} deals with the translation problem when there are significant geometry variations between domains. It disentangles images into the latent appearance and geometry spaces. 
Li~\etal's method~\cite{li2019attribute} can learn multiple domains and multimodal translation simultaneously but requires specific labels to generate style-guided results. 
MSGAN~\cite{mao2019msgan} presents a mode seeking regularization term for increasing the diversity of generated results. 
Recently, some focus on the few-shot and one-shot unsupervised I2I problem~\cite{liu2019few,lin2020tuigan}.
\section{Proposed method}
\label{sec:method}

For style transfer between two image domains $X_A$ and $X_B$, our goal is to find two mappings $\Phi_{X_A\rightarrow{X_B}}$ and $\Phi_{X_B\rightarrow{X_A}}$ for the conversion between images in $X_A$ and $X_B$. 

\subsection{Model overview}
\label{sec:model_overview}
\vspace{\subsecmargin}

\figref{overview} depicts the basic framework of our method. 
In addition to basic content encoders $\{E_A^c, E_B^c\}$, style encoders $\{E_A^s, E_B^s\}$, and generators $\{G_A, G_B\}$, our method also learns the domain-specific mapping functions $\{\Phi_{C\rightarrow{C_A}}, \Phi_{C\rightarrow{C_B}}\}$ which map a feature in the shared domain-invariant content space to the domain-specific content spaces of $X_A$ and $X_B$ respectively.  
For the scenario of $X_A\!\rightarrow\!X_B$, the content image $x_A\!\in\!X_A$ is first encoded into the domain-invariant content feature $c_A = E_A^c(x_A)$, where $c_A \in C^{DI}$. 
Similarly, the style image $x_B\!\in\!X_B$ is encoded into the style feature $s_B = E_B^s(x_B)$, where $s_B\!\in\!S_B$. 
MUNIT and DRIT generate the result simply by $x_{A\rightarrow{B}} = G_B(c_A, s_B)$.
We conjecture that the domain-invariant content feature $c_A$ might not be good enough to go with the style feature $s_B$ to generate a good image in the domain $X_B$. It would be better to align $c_A$ with the content characteristics of the target domain before synthesis.
Thus, for improving the results, our method uses the additional domain-specific content mappings to map the content feature into the domain-specific content spaces. 
In this scenario, the function $\Phi_{C\rightarrow{C_B}}$ is for mapping the content feature $c_A$ to the content space of $X_B$. 
At a high level, it obtains the content feature $c_{A \rightarrow B}$ by aligning the original content in $X_A$ into the content space of the domain $X_B$, probably through semantic correspondences between domains.
Note that, different from previous approaches that directly use the content feature $c_A$ in $X_A$, our method aligns the content feature into the content space of $X_B$, which better matches the style feature and improves results.
The generator $G_B$ then takes the content feature $c_{A \rightarrow B}$ and the style feature $s_B$ for synthesizing the output image $x_{A \rightarrow B}$.
In sum, for the scenario of $X_A\!\rightarrow\!X_B$, our method generates the output $x_{A\rightarrow{B}}$ by
\begin{equation*}
    x_{A\rightarrow{B}} = \Phi_{X_A\rightarrow{X_B}}(x_A, x_B) = G_B(\Phi_{C\rightarrow{C_B}}(c_A), s_B)\,.
\end{equation*}
Similarly, for the scenario of $X_B\!\rightarrow\!X_A$, we have
\begin{equation*}
    x_{B\rightarrow{A}} = \Phi_{X_B\rightarrow{X_A}}(x_B, x_A) = G_A(\Phi_{C\rightarrow{C_A}}(c_B), s_A)\,.
\end{equation*}

\subsection{Learning domain-specific content mappings} 
\label{sec:domain_specific_mapping}
\vspace{\subsecmargin}

The key question is how to learn the domain-specific content mappings $\Phi_{C\rightarrow{C_A}}$ and $\Phi_{C\rightarrow{C_B}}$ in the latent space. Following the design of MUNIT~\cite{huang2018munit}, our content encoder $E_A^{c}$ is composed of two parts, $E_A^{hc}$ and $E_{res}^{hc}$, as shown in \figref{overview}. The first part $E_A^{hc}$ consists of several strided convolution layers for downsampling. The second part, $E_{res}^{hc}$, is composed of several residual blocks for further processing. We choose to share $E_{res}^{hc}$ in the encoders of both domains, but using separate residual blocks for different domains also works.
Thus, we have 
\begin{equation}
\begin{split}
    c_A = E_A^c(x_A) = E_{res}^{hc}( E_A^{hc}(x_A))\,, \ \  c_B = E_B^c(x_B) = E_{res}^{hc}( E_B^{hc}(x_B))\,.
\end{split}
\end{equation}
We opt to use the intermediate content features, $h_A\!\!=\!\!E_A^{hc}(x_A)$ and $h_B\!\!=\!\!E_B^{hc}(x_B)$, as the domain-specific content features. 
In this way, $h_A$ and $h_B$ are domain-specific, and the residual blocks $E_{res}^{hc}$ is responsible for projecting them to the domain-invariant space through minimizing the domain invariant content loss that will be described later in \eqnref{dic_loss}. 
For finding the domain-specific mapping $\Phi_{C\rightarrow{C_A}}$, we require that its output resembles the domain-specific content feature $h_A$ so that it can keep domain-specific properties for $X_A$. Similarly for $\Phi_{C\rightarrow{C_B}}$. Thus, we have the following domain-specific content (dsc) reconstruction loss. By minimizing the loss, we can obtain the mappings. 
\begin{equation}
\begin{split}
    \label{eq:domain_loss}
    L_1^{dsc_A} &= \mathbb{E}_{x_A}[\left\|E_{A}^{hc}(x_A)-\Phi_{C\rightarrow{C_A}}(E_A^c(x_A))\right \|_1]\,, \\
    L_1^{dsc_B} &= \mathbb{E}_{x_B}[\left\|E_{B}^{hc}(x_B)-\Phi_{C\rightarrow{C_B}}(E_B^c(x_B))\right \|_1]\,.
\end{split}
\end{equation}

\subsection{Losses} 
\vspace{\subsecmargin}

In addition to the domain-specific content reconstruction loss introduced in \eqnref{domain_loss}, our formulation consists of several other losses: some for style transfer and the others for image-to-image translation.

\subsubsection{3.3.1 Loss for style transfer.} 
The style reconstruction loss is employed for training the style encoders.

\heading{Style reconstruction loss.} To ensure the style encoders encode meaningful style features, inspired by the Gaussian priors in DRIT~\cite{lee2018drit}, when given a style feature $s_A$ randomly sampled from a Gaussian distribution, we need to reconstruct it back to the original style feature. The loss is defined as the following.
\begin{align}
    \label{eq:style_loss}
    L_{1}^{s_A} = \mathbb{E}_{s_A, c_{B\rightarrow{A}}}[\left\|E_{A}^{s}( G_A(c_{B\rightarrow{A}}, s_A)) )- s_A\right \|_1]\,,
\end{align}
where $c_{B\rightarrow{A}} = \Phi_{C\rightarrow{C_A}}(E_B^c(x_B))$.

\subsubsection{3.3.2 Losses for image-to-image translation} 
Similar to other I2I methods~\cite{huang2018munit,lee2018drit}, we adopt adversarial, image reconstruction, domain-invariant content reconstruction, and cycle-consistency losses to facilitate training. 

\heading{Domain-invariant content loss.} Even though we employ domain-specific content features to learn better alignment between domains, we still need the domain-invariant content space for cross-domain translation. We require that $x_A$ and $x_{A\rightarrow{B}}$ in \figref{overview} have the same domain-invariant content feature, \ie,
\begin{align}
    L_{1}^{dic_A} = \mathbb{E}_{x_A, x_{A\rightarrow{B}}}[\left\|E_{B}^c(x_{A\rightarrow{B}})) - E_{A}^c(x_A)\right \|_1]\,.
    \label{eq:dic_loss}
\end{align}

\heading{Image reconstruction loss.} Just like variational autoencoders (VAEs)~\cite{kingma2013vae}, the image reconstruction loss is used to make sure the generator can reconstruct the original image within a domain.
\begin{align}
L_{1}^{x_A}\!=\!\mathbb{E}_{x_A}[\left\| G_A(\Phi_{C\rightarrow{C_A}}(E_A^c(x_A)), E_A^s(x_A))\!-\!x_A\right \|_1]\,.
\end{align}
Note the original content feature $c_A$ is domain-invariant, we need to map $c_A$ to the domain-specific content feature $c_{A\rightarrow{A}}$. Taking the content feature $c_{A\rightarrow{A}}$ and the style feature $s_A$, the generator $G_A$ should reconstruct the original image $x_A$.

\heading{Adversarial loss.} Like MUNIT~\cite{huang2018munit}, we employ the adversarial loss of LSGAN~\cite{mao2017lsgan} to minimize the discrepancy between the distributions of the real images and the generated images.
\begin{align}
    L_{D_{adv}}^A &= \tfrac{1}{2} \mathbb{E}_{x_A}[((D_A(x_A)-1))^2] + \tfrac{1}{2} \mathbb{E}_{x_{B\rightarrow{A}}}[(D_A(x_{B\rightarrow{A}}))^2] \nonumber \\
    L_{G_{adv}}^A &= \tfrac{1}{2} \mathbb{E}_{x_{B\rightarrow{A}}}[(D_A(x_{B\rightarrow{A}}) -1)^2]\,,
\end{align}
where $D_A$ is the discriminator of domain $X_A$.

\heading{Cycle-consistency loss.} The cycle consistency constraint was proposed in CycleGAN~\cite{zhu2017cyclegan} and has been proved useful in unsupervised I2I translation. When a given input $x_A$ passes through the cross-domain translation pipeline $A\!\rightarrow\!{B}\!\rightarrow\!{A}$, it should be reconstructed back to $x_A$ itself, \ie,
\begin{align}
L_{cc}^{x_A} = \mathbb{E}_{x_A}[\left\| x_{A\rightarrow{B}\rightarrow{A}}-x_A\right \|_1]\,.
\end{align}

\heading{3.3.3 Total loss.} Our goal is to perform the cross-domain training, so we need to train on both directions. Losses on the other direction are defined similarly. We combine each pair of dual terms together such as $L_{cc}^x = L_{cc}^{x_A}+L_{cc}^{x_B}$, $L_{1}^x = L_{1}^{x_A}+L_{1}^{x_B}$ \etc. Finally, the total loss is defined as
\begin{align}
L_{G_{total}} &= \lambda_{cc}L_{cc}^x + \lambda_{x}L_{1}^x + \lambda_{dsc}L_{1}^{dsc} + \lambda_{dic}L_{1}^{dic} + \lambda_{s}L_{1}^s + \lambda_{adv} L_{G_{adv}} \nonumber\\ 
L_{D} &= \lambda_{adv} L_{D_{adv}}
\end{align}
where $\lambda_{cc}$, $\lambda_{x}$, $\ldots$ are hyper-parameters for striking proper balance among losses.

\section{Experiments}
\label{sec:exp}

This section first introduces the datasets used for evaluation and competing methods. Then, qualitative and quantitative comparisons are reported, followed by discussions.

\subsection{Datasets}
\vspace{\subsecmargin}
We present results on more challenging cross-domain translation problems in the paper. More results can be found in the supplementary document.

\heading{\itoi{Photo} $\!\rightleftarrows\!$ \itoi{Monet}}~\cite{zhu2017cyclegan}. This dataset is provided by CycleGAN, which includes real scenic images and Monet-style paintings.

\heading{\itoi{Cat} $\!\rightleftarrows\!$ \itoi{Dog}}~\cite{lee2018drit}. DRIT collects 771 cat images and 1,264 dog images. There is a significant gap between these two domains, and it is necessary to take the semantic information into account for translation. 

\heading{\itoi{Photograph} $\!\rightleftarrows\!$ \itoi{Portrait}}~\cite{lee2018drit}. This dataset is also provided by DRIT, which includes portraits and photographs of human faces. The gap is smaller than the \itoi{Cat} $\!\rightleftarrows\!$ \itoi{Dog} dataset. However, the difficulty of the translation is to preserve identity while performing style translation.

\subsection{Competing methods} 
\vspace{\subsecmargin}
We compare our method with three image-to-image translation methods and three style transfer methods.

\heading{Image-to-Image translation methods.} MUNIT~\cite{huang2018munit} disentangles the images into the domain-invariant content space and domain-specific style spaces. 
MSGAN~\cite{mao2019msgan} extends DRIT~\cite{lee2018drit} by adopting the mode seeking regularization term to improve the diversity of the generated images. Since MSGAN generally generates better results than DRIT, we do not compare with DRIT. We use the pre-trained \itoi{cat $\!\rightleftarrows\!$ dog} model provided by MSGAN while training MSGAN on the other two datasets. 
GDWCT~\cite{cho2019gdwct} applies WCT~\cite{li2017universal} to I2I translation and obtains better translation results.
For the I2I translation methods: MUNIT, MSGAN, GDWCT and ours, we train them on NVIDIA GTX 1080Ti for two days. 

\heading{Style transfer methods.}
AdaIN~\cite{huang2017adain} uses the adaptive instance normalization layer to match the mean and variance of content features to those of style features for style transfer. 
Liao~\etal~\cite{liao2017semantic} assume that the content and style images have perceptually similar semantic structures. Their method first extracts features using DNN, then adopts a coarse-to-fine strategy for computing the nearest-neighbor field and matching features.
Luan~\etal~\cite{luan2017deep} apply the segmentation masks to segment the semantic of content and style images and employ iterative optimization to calculate the loss functions. 
Note it is not completely fair to compare  I2I methods with the style transfer methods as some of them have different definitions of styles and could use less information than I2I methods. 

\begin{figure*}[!t]
  \centering
  \includegraphics[width=\textwidth]{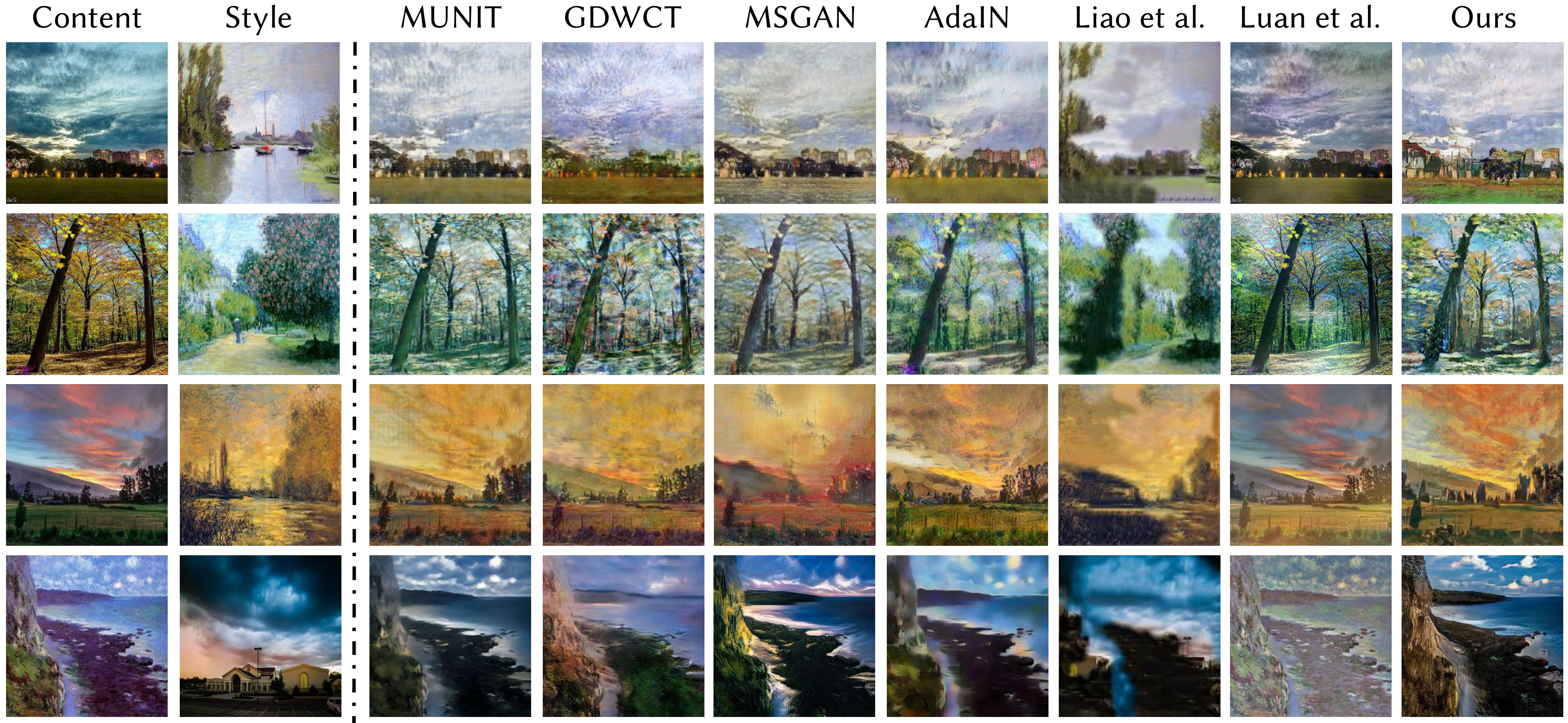}
  \caption{{\bf Comparisons on {Photo}$\rightleftarrows${Monet}.} 
  }
  \label{fig:exp_monet}
   \vspace{.4\figmargin}
\end{figure*}

\begin{figure*}[!t]
  \centering
  \includegraphics[width=\textwidth]{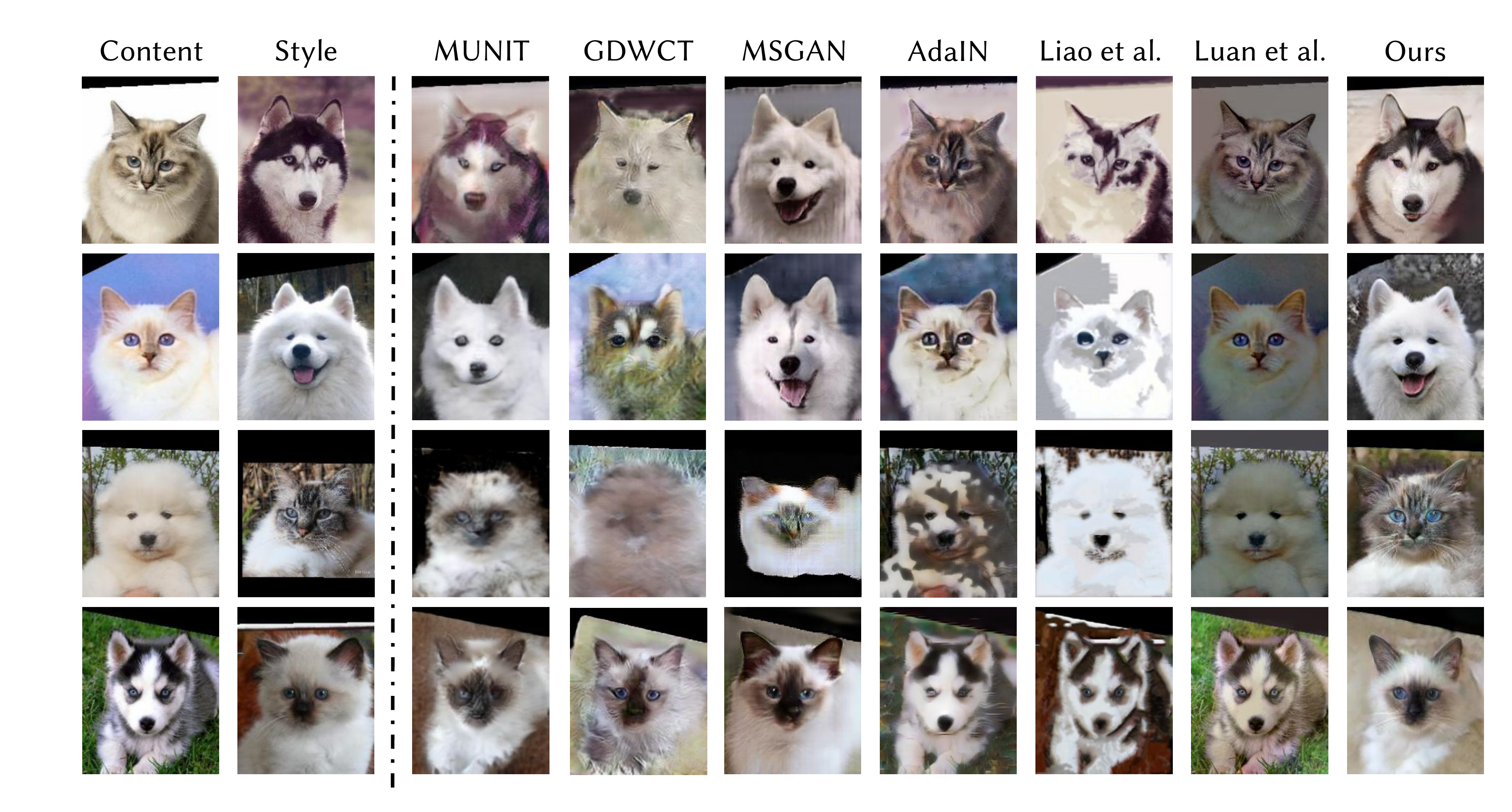}
  \caption{{\bf Comparisons on Cat$\rightleftarrows$Dog.}}
  \label{fig:exp_c2g}
   \vspace{\figmargin}
\end{figure*}

\begin{figure*}[!t]
  \centering
  \includegraphics[width=\textwidth]{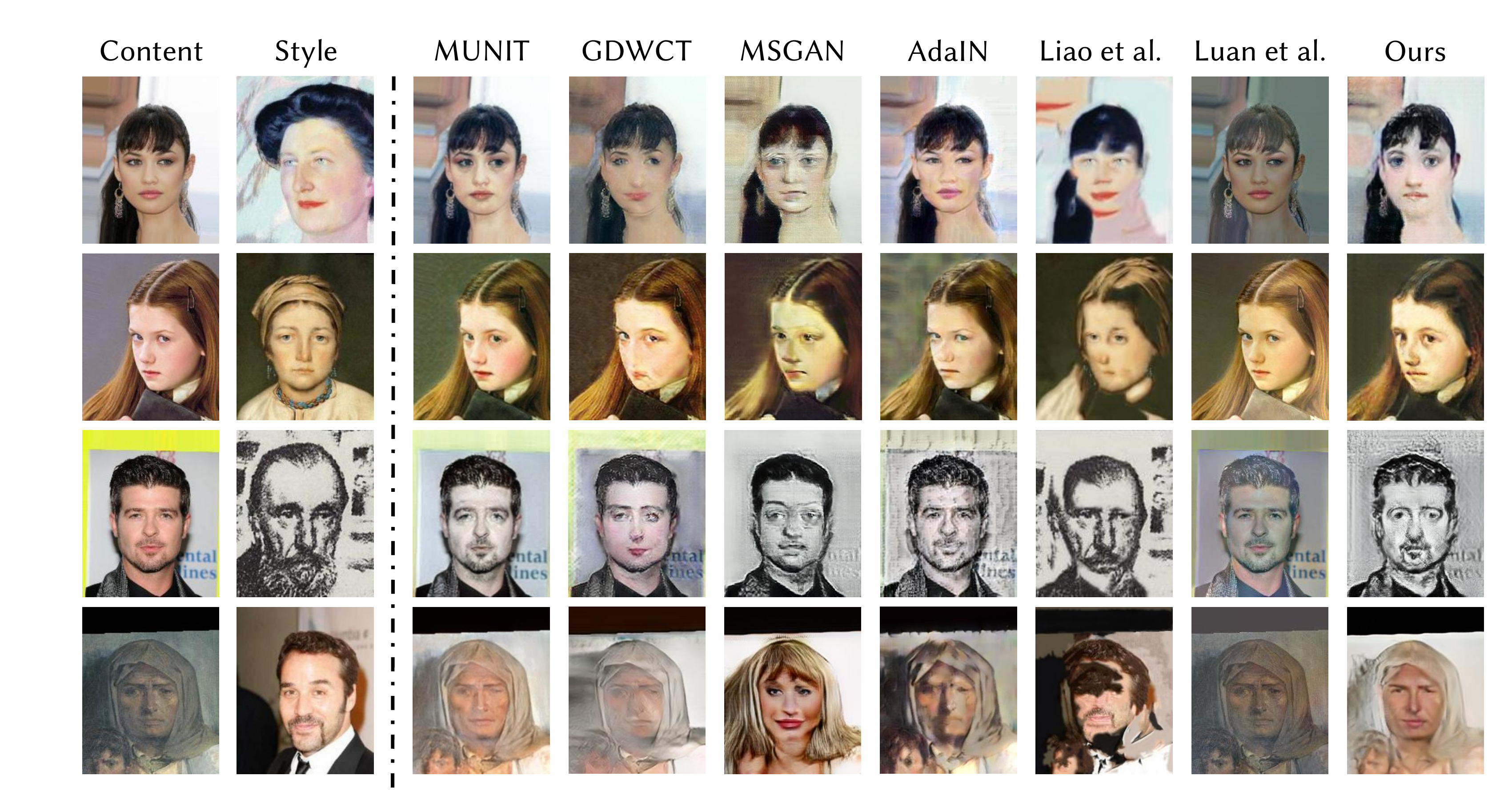}
  \caption{{\bf Comparisons on Photograph$\rightleftarrows$Portrait.}
  }
  \label{fig:exp_i2p}
   \vspace{\figmargin}
\end{figure*}

\subsection{Qualitative comparisons}

\figref{exp_monet} shows results of \itoi{photo $\!\rightleftarrows\!$ Monet}. Since this is a simpler scenario, most methods give reasonable results. However, our method still generates results of higher quality than other methods. 
\figref{exp_c2g} presents results for the \itoi{cat $\!\rightleftarrows\!$ dog} task. The results of MUNIT and GDWCT have the same problem that the characteristics of the species are not clear. It is not easy to judge the species depicted by the images. MSGAN generates images with more obvious characteristics of the target species. However, it does not preserve the content information as well as our method. In their results, the poses and locations of facial/body features are not necessarily similar to the content images. Our method generates much clearer results that better exhibit the characteristics of target species and preserve layouts of the content images. The style transfer methods~\cite{huang2017adain,liao2017semantic,luan2017deep} have poor performance due to the different assumption of styles and the use of less information. 

\figref{exp_i2p} shows the comparisons of the \itoi{photographs} $\!\rightleftarrows\!$ \itoi{portraits} task. This task is challenging because the identity in the content image must be preserved, thus often requiring better semantic alignment. 
MUNIT preserves the identity very well but does not transfer the style well. On the contrary, GDWCT transfers styles better, but the identity is often not maintained well. MSGAN transfers the style much better than MUNIT and GDWCT but does not perform well on identity preservation. Our method performs both style transfer and identity preservation well. Style transfer methods again provide less satisfactory results than I2I methods in general.
Note that there is subtle expression change in the portraits' expressions in the first two rows of \figref{exp_i2p}. As other disentangled representations, our method does not impose any high-level constraints or knowledge on the decoupling of the style and content features. Thus, the division between the style and content has to be learned from data alone. There could be ambiguity in the division. Other methods except for MUNIT, exhibit similar or even aggravated expression change. MUNIT is an exception as it tends to preserve content but does little on style transfer.

\subsection{Quantitative comparison} 
\vspace{\subsecmargin}
As shown in the previous section, the compared style transfer methods cannot perform cross-domain style transfer well. Thus, we only include image-to-image translation methods in the quantitative comparison. 

\begin{table}[t]
\tabcolsep=7pt
\renewcommand\arraystretch{1.0}
\centering
\caption{{\bf Quantitative comparison.} We use the FID score (lower is better) and the LPIPS score (higher is better) to evaluate the quality and diversity of each method on six types of translation tasks. Red texts indicate the best and blue texts indicate the second best method for each task and metric.}
\label{tab:fid_lpips}
\resizebox{\columnwidth}{!}{
\begin{tabular}{lcccccccc}
\toprule
& \multicolumn{4}{c}{FID$\downarrow$}
& \multicolumn{4}{c}{LPIPS$\uparrow$}          \\ 
\cmidrule(r){2-5}\cmidrule(r){6-9}

& MUNIT & GDWCT & MSGAN & Ours
& MUNIT & GDWCT & MSGAN & Ours \\ \midrule

Cat $\rightarrow$ Dog        
& ${38.09}$  
& ${91.40}$  
& $\second{20.80}$ 
& $\best{13.60}$ 

& ${0.3501}$ 
& ${0.1804}$ 
& $\best{0.5051}$
& $\second{0.4149}$  \\

Dog $\rightarrow$ Cat        
& ${39.71}$  
& ${59.72}$  
& ${\second{28.30}}$ 
& ${\best{19.69}}$  

& ${0.3167}$
& ${0.1573}$
& ${\best{0.4334}}$
& ${\second{0.3174}}$   \\

Monet $\rightarrow$ Photo
& ${\second{85.06}}$  
& ${113.16}$ 
& ${86.72}$  
& ${\best{81.61}}$  

& $\second{0.4282}$
& ${0.2478}$
& ${0.4229}$
& ${\best{0.5379}}$ \\

Photo $\rightarrow$ Monet
& ${77.85}$ 
& ${\second{71.68}}$ 
& ${80.37}$ 
& ${\best{63.94}}$ 

& ${0.4128}$
& ${0.2097}$
& ${\second{0.4306}}$
& ${\best{0.4340}}$ \\

Portrait $\rightarrow$ Photograph 
& ${93.45}$  
& ${83.69}$  
& ${\best{57.07}}$ 
& ${\second{62.44}}$ 

& ${0.1819}$ 
& ${0.1563}$ 
& ${\second{0.3061}}$ 
& ${\best{0.3160}}$ \\

Photograph $\rightarrow$ Portrait 
& ${89.97}$ 
& ${75.86}$ 
& ${\second{57.84}}$
& ${\best{45.81}}$ 

& ${0.1929}$ 
& ${0.1785}$ 
& ${\second{0.2917}}$ 
& ${\best{0.3699}}$ \\ \midrule

Average                     
& $70.69$   
& $82.59$   
& ${\second{55.18}}$
& ${\best{47.85}}$

& $0.3131$ 
& $0.1881$
& ${\second{0.3978}}$
& ${\best{0.3980}}$ \\ \bottomrule

\end{tabular}}
\vspace{\tabmargin}
\end{table}
 
\heading{Quality of images.}
We use the FID score~\cite{heusel2017fid} to measure the similarity between distributions of generated images and real images in the cross-domain translation task. FID is calculated by computing the Fréchet distance through the features extracted from the Inception network. The lower FID score indicates a better quality, and the generated images are closer to the target domain.

We randomly sample 100 test images and generate ten different example-guided results for each image. These results are then used to calculate the FID score for each method. We repeat ten times and report the average scores. As shown in~\tabref{fid_lpips}, our method achieves the best scores except for the task of {Portrait}$\to${Photograph}. Note that it is often more challenging to generate realistic photographs. Thus, the FID scores for the tasks generating photographs are generally worse.  
For other scenarios, our method often has a significantly lower score than other methods. 

\heading{Diversity.}
To measure the diversity among the generated images, we report the LPIPS score~\cite{zhang2018lpips}, which measures feature distances between paired outputs. The higher LPIPS scores indicate better diversity among generated images. We randomly sample 100 content images, and for each of them, we generate 15 paired results. Again, we repeat ten times and report the average scores.  As shown in~\tabref{fid_lpips}, even if our mapping function is not designed to increase diversity, our method achieves good diversity and performs very well for the \itoi{photographs} $\!\rightleftarrows\!$ \itoi{portraits} and \itoi{Monet} $\!\rightleftarrows\!$ \itoi{photos} tasks. MSGAN~\cite{mao2019msgan} specifically adds loss function for promoting the diversity of generated images. Thus, its results also have good diversity, as seen in the \itoi{cat} $\!\rightleftarrows\!$ \itoi{dog} tasks.

\begin{figure}[!t]
  \centering
  \includegraphics[width=\columnwidth]{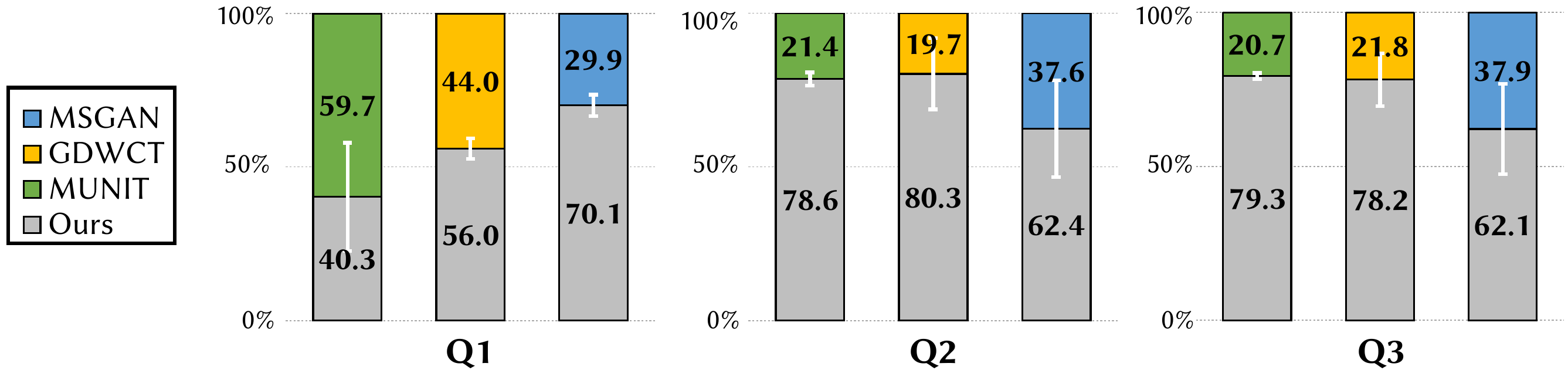}
  \caption{{\bf Result of the user study.} The numbers indicate the percentage of users preferring in the pairwise comparison. We conduct the user study on {Cat}$\to${Dog}, {Dog}$\to${Cat} and {Photograph}$\to${Portrait} translation tasks and report their averages. The white error bar indicates the standard deviation. The complete results can be found in the supplementary.}
  \label{fig:user}
   \vspace{\figmargin}
\end{figure}

\heading{User study.}
For each test set, users are presented with the content image (domain A), the style image (domain B), and two result images, one generated by our method and the other by one of the three compared methods. The result images are presented in random order. 
The users need to select the better image among the two given results for the following three questions.
\begin{itemize}
\small
\setlength{\itemsep}{0pt}
\setlength{\parsep}{0pt}
\setlength{\parskip}{0pt}
\item  \emph{Q1: Which one preserves content information (identity, shape, semantic) better?}
\item  \emph{Q2: Which one performs better style translation (in terms of color, pattern)?}
\item  \emph{Q3: Which one is more likely to be a member of the domain B?}
\end{itemize}
As shown in~\figref{user}, MUNIT preserves visual characteristics of the content very well but does very little on transferring styles. Thus, MUNIT has high scores in \textit{Q1} while having very low scores in \textit{Q2} and \textit{Q3}. GDWCT has a similar performance to MUNIT.
On the contrary, MSGAN is less capable of preserving content while performing better in style transfer than MUNIT/GDWCT. For style transfer (\textit{Q2} and \textit{Q3}), MSGAN performs slightly better than our method for the scenario of \itoi{cats} $\!\rightarrow\!$ \itoi{dogs}, but falls significantly behind our method in other scenarios. For content preservation (\textit{Q1}), MSGAN is much worse than ours. 

\begin{figure}[t]
  \centering
  \includegraphics[width=.98\columnwidth]{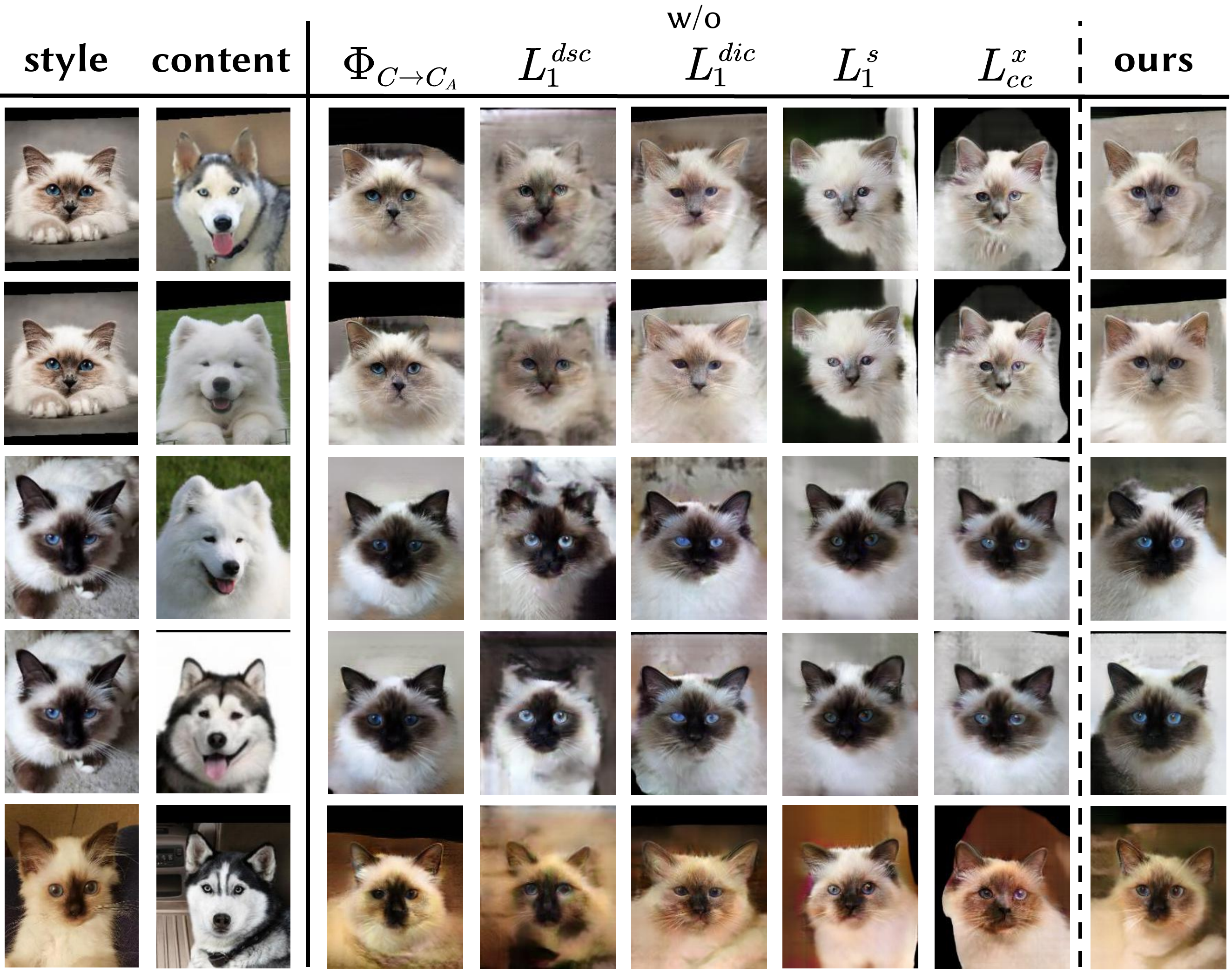}
  \caption{{\bf Visual comparisons of the ablation study.} We show the results without (w/o) the domain-specific content mapping $\Phi_{C\rightarrow{C_A}}$, domain-specific content loss $L_1^{dsc}$, domain-invariant content loss $L_1^{dic}$, style reconstruction loss $L_1^s$, and cycle-consistency loss $L_{cc}^x$ for several examples of {Dog}$\to${Cat}.}
  \label{fig:domain_ablation}
   \vspace{\figmargin}
\end{figure}

\subsection{Discussions}

\heading{{Ablation studies.}} \figref{domain_ablation} gives the results without the individual losses in our method for better understanding their utility. We show the ablation study on {Dog}$\to${Cat} since this task is more challenging.
Without the domain-specific content mapping, the spatial layouts of the content images can not be preserved well. With the mapping, poses and sizes of the synthesized cats better resemble those of the dogs in the content images.  
The proposed loss $L_1^{dsc}$ ensures the remapped feature resembles the domain-specific feature $h_A$, and it is essential to the learning of $\Phi_{C\rightarrow{C_A}}$. Our model cannot learn the correct mapping without $L_1^{dsc}$. Without $L_1^{dic}$, content preservation is less stable. 
The model can not learn the correct style without $L_1^s$ because there is no cue to guide proper style encoding. The cycle consistency loss $L_{cc}^x$ is essential for unsupervised I2I learning. Without it, content and style cannot be learned in an unsupervised manner. 

\begin{figure}[t]
    \centering 
    \begin{subfigure}[t]{.49\columnwidth} 
        \includegraphics[width=\columnwidth]{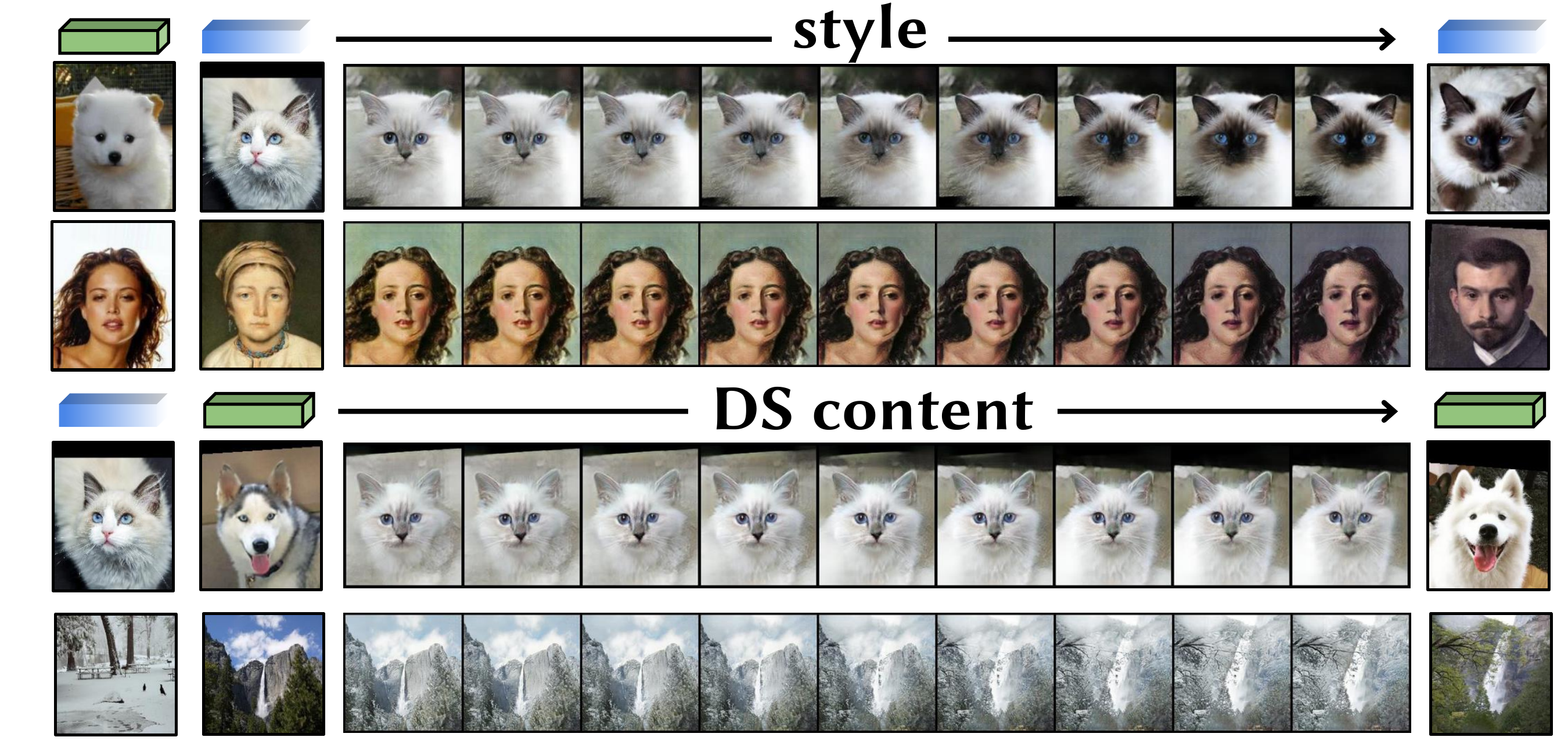}
        \caption{interpolation of style and domain-specific content vectors}
    \end{subfigure}
    \begin{subfigure}[t]{.49\columnwidth} 
        \includegraphics[width=.99\columnwidth]{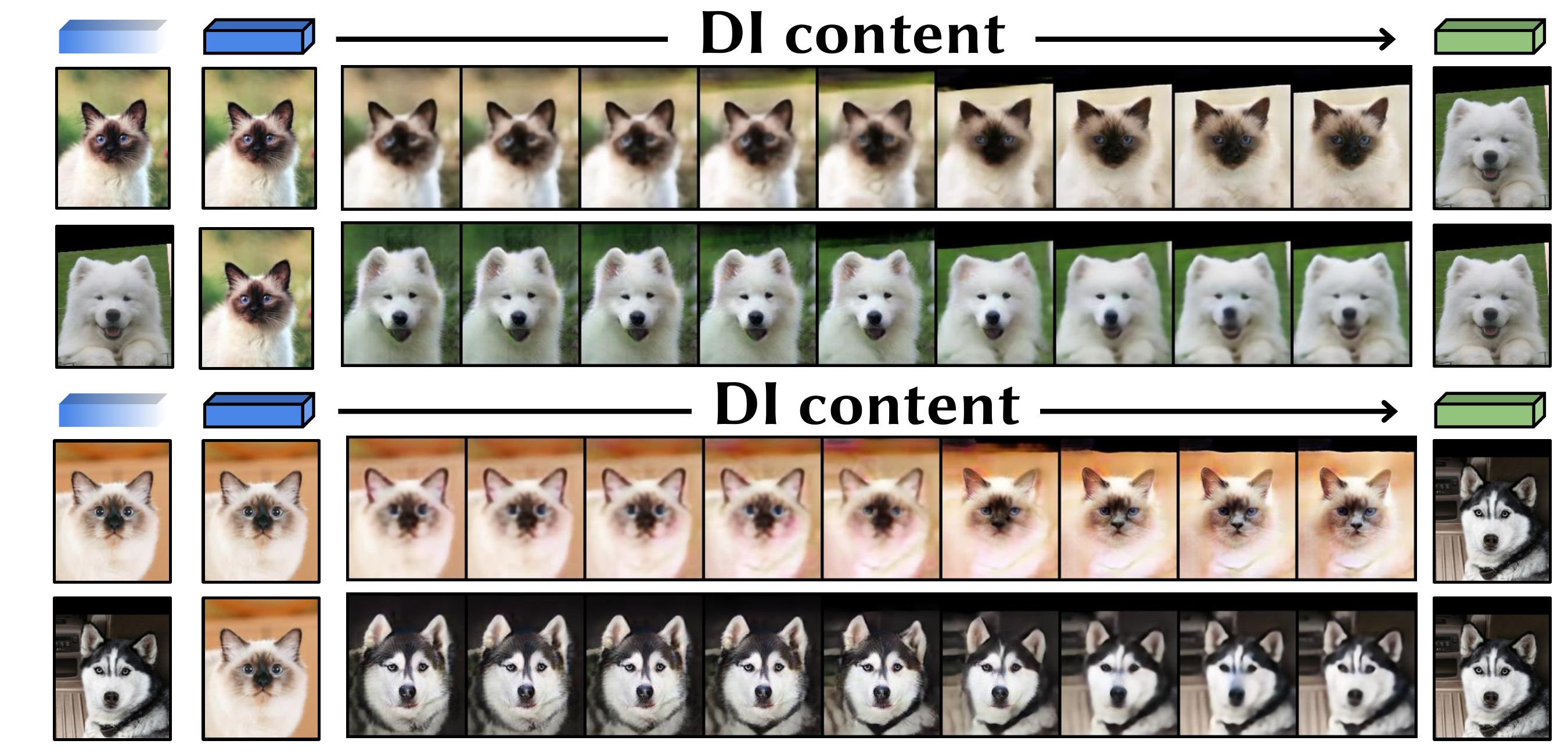}
         \caption{interpolation of domain-invariant content vectors}
    \end{subfigure}  
        \caption{{\bf Interpolation in the latent space.}} 
        \label{fig:exp_inter}
\end{figure}

\heading{Interpolation in the latent space.}
\figref{exp_inter}(a) shows the results of style interpolation for the fixed content at the top and domain-specific content interpolation for the fixed style at the bottom. They demonstrate that our model has nice continuity property in the latent space.

Note that our content space is domain-specific, and thus the content interpolation must be performed in the same domain. It seems that we cannot perform content interpolation between a cat and a dog since their content vectors are in different spaces. 
However, it is still possible to perform content interpolation between two domains by interpolating in the shared space and then employing the domain-specific mapping. Taking the first row of \figref{exp_inter}(b) as an example, the first content image is a cat (domain A) at one pose while the second content is a dog (domain B) at another pose. We first obtain their content vectors in the shared content space and then perform interpolation in the shared space. Next, since the style is a cat (domain A), we employ the mapping $\Phi_{C\rightarrow{C_A}}$ to remap the interpolated content vector into the cat's domain-specific content space. By combining it with the cat's style vector, we obtain the cat's image at the interpolated pose between the cat's and dog's poses. \figref{exp_inter}(b) shows several examples of linear interpolation between content images from two different domains. The results show that the domain-specific mapping helps align the content feature with the target style.

\begin{figure}[t]
    \centering
    \begin{subfigure}{.49\columnwidth} 
        \includegraphics[width=\columnwidth]{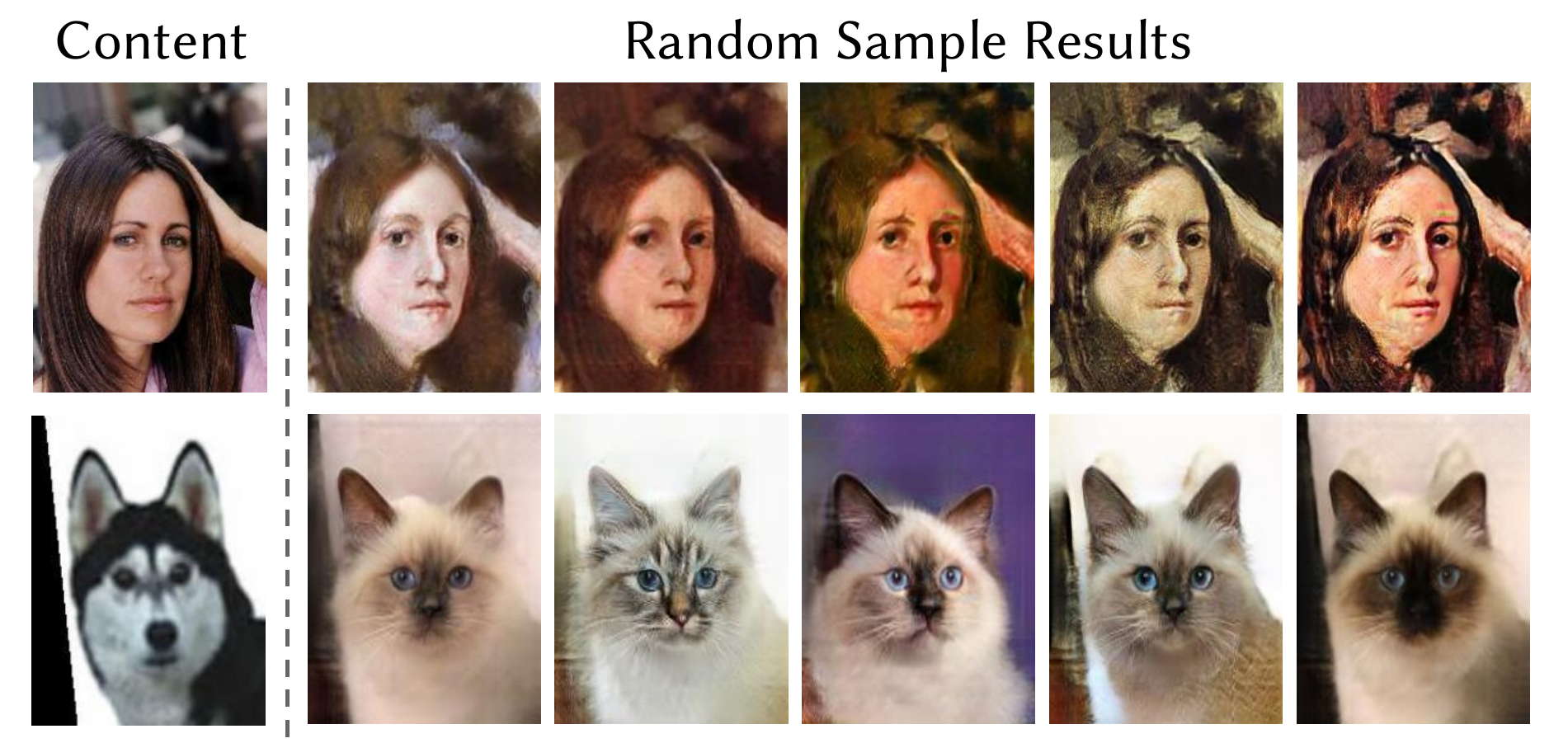}
        \caption{multi-modal results}
    \end{subfigure}
    \begin{subfigure}{.49\columnwidth} 
        \includegraphics[width=\columnwidth]{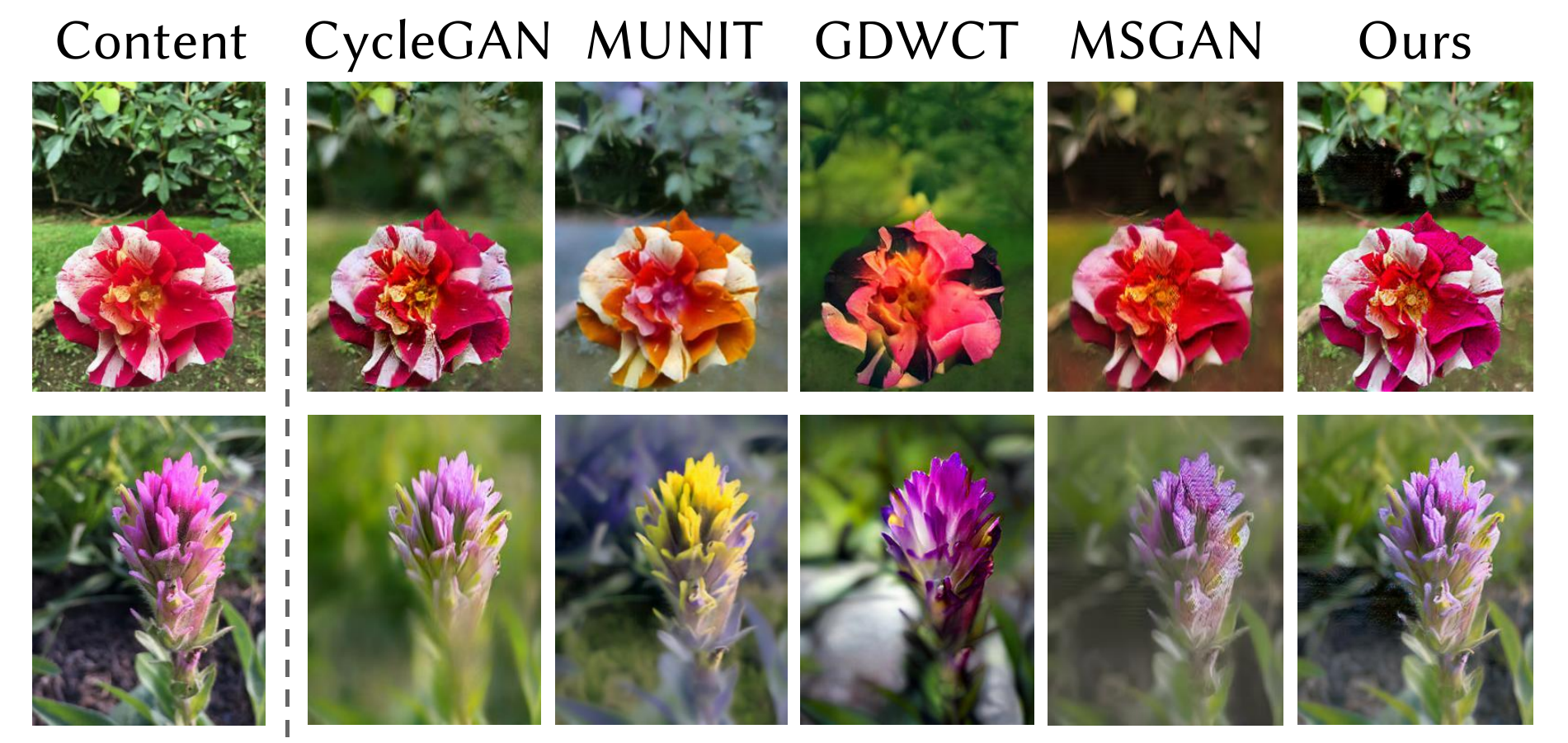}
        \caption{{iPhone}$\to${DSLR}}
    \end{subfigure}
        \caption{{\bf Additional results.} (a) multi-modal results and (b) {iPhone}$\to${DSLR}.
        } 
        \label{fig:dis_applications}
\end{figure}

\heading{Other tasks.} 
\figref{dis_applications}(a) shows the multi-modal results of our method by combining a fixed content vector with several randomly sampled style vectors. 
\figref{dis_applications}(b) demonstrates the results for 
the \itoi{Iphone} $\!\rightleftarrows\!$ \itoi{DSLR} task provided by CycleGAN~\cite{zhu2017cyclegan}. The task is easy for CycleGAN because it is a task of one-to-one mapping and does not involve the notion of style features. However, I2I methods could run into problems with color shifting. Our results have better color fidelity than other I2I methods while successfully generating the shallow depth-of-field effects. 
 
\begin{figure}[!t]
    \centering  
    \includegraphics[width=1.0\columnwidth]{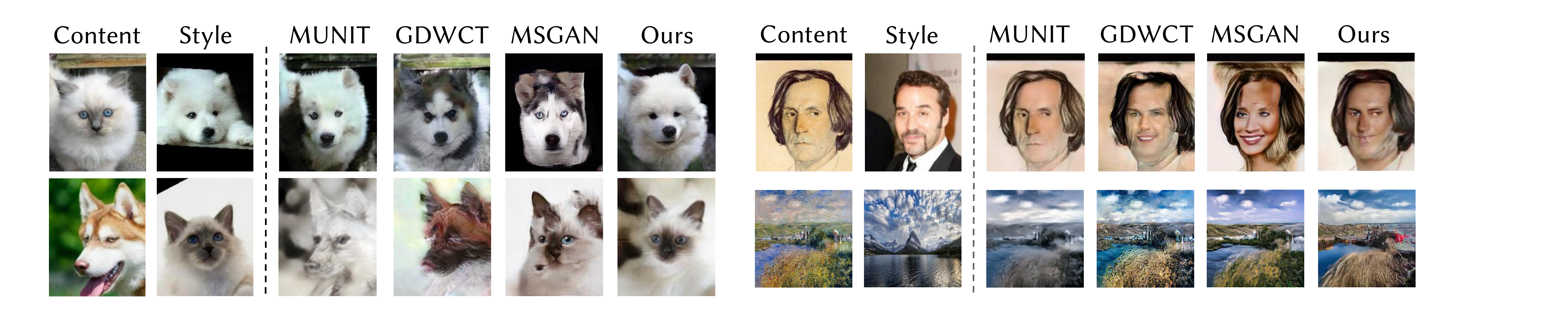}
    \caption{{\bf Failure examples.}
    }
    \label{fig:dis_fail}
\end{figure}

\heading{Failure cases.} 
\figref{dis_fail} gives examples in which our method is less successful. For the cases on the left, the poses are rare in the training set, particularly the one on the bottom. Thus, the content is not preserved as well as other examples.  For the examples on the right, the target domains are photographs. They are more challenging, and our method could generate less realistic images. Even though our approach does not produce satisfactory results for these examples, our results are still much better than those of other methods.

\section{Conclusion}

This paper proposes to use domain-specific content mappings to improve the quality of the image-to-image translation with the disentangled representation. By aligning the content feature into the domain-specific content space, the disentangled representation becomes more effective. Experiments on style transfer show that the proposed method can better handle more challenging translation problems, which would require more accurate semantic correspondences. In the future, we would like to explore the possibility of applying the domain-specific mapping to other I2I translation frameworks.

\vspace{1mm}
\heading{Acknowledgments.} This work was supported in part by MOST under grant 107-2221-E-002-147-MY3 and MOST Joint Research Center for AI Technology and All Vista Healthcare under grant 109-2634-F-002-032.

\bibliographystyle{splncs04}
\bibliography{ref}

\begin{thebibliography}{10}
\providecommand{\url}[1]{\texttt{#1}}
\providecommand{\urlprefix}{URL }
\providecommand{\doi}[1]{https://doi.org/#1}

\bibitem{chen2017stylebank}
Chen, D., Yuan, L., Liao, J., Yu, N., Hua, G.: {StyleBank}: An explicit
  representation for neural image style transfer. In: \CVPR (2017)

\bibitem{cho2019gdwct}
Cho, W., Choi, S., Keetae~Park, D., Shin, I., Choo, J.: Image-to-image
  translation via group-wise deep whitening-and-coloring transformation. In:
  \CVPR (2019)

\bibitem{choi2018stargan}
Choi, Y., Choi, M., Kim, M., Ha, J.W., Kim, S., Choo, J.: {StarGAN}: Unified
  generative adversarial networks for multi-domain image-to-image translation.
  In: \CVPR (2018)

\bibitem{gatys2015texture}
Gatys, L., Ecker, A.S., Bethge, M.: Texture synthesis using convolutional
  neural networks. In: \NIPS (2015)

\bibitem{gatys2016image}
Gatys, L.A., Ecker, A.S., Bethge, M.: Image style transfer using convolutional
  neural networks. In: \CVPR (2016)

\bibitem{heusel2017fid}
Heusel, M., Ramsauer, H., Unterthiner, T., Nessler, B., Hochreiter, S.: {GANs}
  trained by a two time-scale update rule converge to a local nash equilibrium.
  In: \NIPS (2017)

\bibitem{huang2017adain}
Huang, X., Belongie, S.: Arbitrary style transfer in real-time with adaptive
  instance normalization. In: \ICCV (2017)

\bibitem{huang2018munit}
Huang, X., Liu, M.Y., Belongie, S., Kautz, J.: Multimodal unsupervised
  image-to-image translation. In: \ECCV (2018)

\bibitem{isola2017pix2pix}
Isola, P., Zhu, J.Y., Zhou, T., Efros, A.A.: Image-to-image translation with
  conditional adversarial networks. In: \CVPR (2017)

\bibitem{johnson2016perceptual}
Johnson, J., Alahi, A., Fei-Fei, L.: Perceptual losses for real-time style
  transfer and super-resolution. In: \ECCV (2016)

\bibitem{Kim2019UGATIT}
Kim, J., Kim, M., Kang, H., Lee, K.: {U-GAT-IT:} unsupervised generative
  attentional networks with adaptive layer-instance normalization for
  image-to-image translation. arXiv preprint arXiv:1907.10830  (2019)

\bibitem{kingma2013vae}
Kingma, D.P., Welling, M.: Auto-encoding variational bayes. arXiv preprint
  arXiv:1312.6114  (2013)

\bibitem{lee2018drit}
Lee, H.Y., Tseng, H.Y., Huang, J.B., Singh, M., Yang, M.H.: Diverse
  image-to-image translation via disentangled representations. In: \ECCV (2018)

\bibitem{li2019attribute}
Li, X., Hu, J., Zhang, S., Hong, X., Ye, Q., Wu, C., Ji, R.: Attribute guided
  unpaired image-to-image translation with semi-supervised learning. arXiv
  preprint arXiv:1904.12428  (2019)

\bibitem{li2018learning}
Li, X., Liu, S., Kautz, J., Yang, M.H.: Learning linear transformations for
  fast arbitrary style transfer. arXiv preprint arXiv:1808.04537  (2018)

\bibitem{li2017universal}
Li, Y., Fang, C., Yang, J., Wang, Z., Lu, X., Yang, M.H.: Universal style
  transfer via feature transforms. In: \NIPS (2017)

\bibitem{liao2017semantic}
Liao, J., Yao, Y., Yuan, L., Hua, G., Kang, S.B.: Visual attribute transfer
  through deep image analogy. In: \SIGGRAPH (2017)

\bibitem{lin2020tuigan}
Lin, J., Pang, Y., Xia, Y., Chen, Z., Luo, J.: {TuiGAN}: Learning versatile
  image-to-image translation with two unpaired images. \ECCV  (2020)

\bibitem{liu2017unsupervised}
Liu, M.Y., Breuel, T., Kautz, J.: Unsupervised image-to-image translation
  networks. In: \NIPS (2017)

\bibitem{liu2019few}
Liu, M.Y., Huang, X., Mallya, A., Karras, T., Aila, T., Lehtinen, J., Kautz,
  J.: Few-shot unsupervised image-to-image translation. In: \ICCV (2019)

\bibitem{lu2017semantic}
Lu, M., Zhao, H., Yao, A., Xu, F., Chen, Y., Zhang, L.: Decoder network over
  lightweight reconstructed feature for fast semantic style transfer. In: \ICCV
  (2017)

\bibitem{luan2017deep}
Luan, F., Paris, S., Shechtman, E., Bala, K.: Deep photo style transfer. In:
  \CVPR (2017)

\bibitem{mao2019msgan}
Mao, Q., Lee, H.Y., Tseng, H.Y., Ma, S., Yang, M.H.: Mode seeking generative
  adversarial networks for diverse image synthesis. In: \CVPR (2019)

\bibitem{mao2017lsgan}
Mao, X., Li, Q., Xie, H., Lau, R.Y., Wang, Z., Paul~Smolley, S.: Least squares
  generative adversarial networks. In: \ICCV (2017)

\bibitem{mejjati2018unsupervised}
Mejjati, Y.A., Richardt, C., Tompkin, J., Cosker, D., Kim, K.I.: Unsupervised
  attention-guided image-to-image translation. In: \NIPS (2018)

\bibitem{mo2018instagan}
Mo, S., Cho, M., Shin, J.: {InstaGAN}: Instance-aware image-to-image
  translation. arXiv preprint arXiv:1812.10889  (2018)

\bibitem{pumarola2018ganimation}
Pumarola, A., Agudo, A., Martinez, A.M., Sanfeliu, A., Moreno-Noguer, F.:
  {GANimation}: Anatomically-aware facial animation from a single image. In:
  \ECCV (2018)

\bibitem{wu2019transgaga}
Wu, W., Cao, K., Li, C., Qian, C., Loy, C.C.: {TransGaGa}: Geometry-aware
  unsupervised image-to-image translation. arXiv preprint arXiv:1904.09571
  (2019)

\bibitem{xiao2018elegant}
Xiao, T., Hong, J., Ma, J.: Elegant: Exchanging latent encodings with {GAN} for
  transferring multiple face attributes. In: \ECCV (2018)

\bibitem{yi2017dualgan}
Yi, Z., Zhang, H., Tan, P., Gong, M.: {DualGAN}: Unsupervised dual learning for
  image-to-image translation. In: \ICCV (2017)

\bibitem{zhang2018lpips}
Zhang, R., Isola, P., Efros, A.A., Shechtman, E., Wang, O.: The unreasonable
  effectiveness of deep features as a perceptual metric. In: \CVPR (2018)

\bibitem{zheng2019multibranch}
Zheng, Z., Yu, Z., Zheng, H., Wu, Y., Zheng, B., Lin, P.: Generative
  adversarial network with multi-branch discriminator for cross-species
  image-to-image translation. arXiv preprint arXiv:1901.10895  (2019)

\bibitem{zhu2017cyclegan}
Zhu, J.Y., Park, T., Isola, P., Efros, A.A.: Unpaired image-to-image
  translation using cycle-consistent adversarial networks. In: \ICCV (2017)

\bibitem{zhu2017bicyclegan}
Zhu, J.Y., Zhang, R., Pathak, D., Darrell, T., Efros, A.A., Wang, O.,
  Shechtman, E.: Toward multimodal image-to-image translation. In: \NIPS (2017)

\end{thebibliography}

\end{document}